# Constraint Processing in Lifted Probabilistic Inference

**Jacek Kisyński and David Poole**
Department of Computer Science
University of British Columbia
Vancouver, B.C. V6T 1Z4
{kisynski, poole}@cs.ubc.ca

## Abstract

First-order probabilistic models combine representational power of first-order logic with graphical models. There is an ongoing effort to design lifted inference algorithms for first-order probabilistic models. We analyze lifted inference from the perspective of constraint processing and, through this viewpoint, we analyze and compare existing approaches and expose their advantages and limitations. Our theoretical results show that the wrong choice of constraint processing method can lead to exponential increase in computational complexity. Our empirical tests confirm the importance of constraint processing in lifted inference. This is the first theoretical and empirical study of constraint processing in lifted inference.

## 1　INTRODUCTION

Representations that mix graphical models and first-order logic—called either first-order or relational probabilistic models—were proposed nearly twenty years ago (Breese, 1992; Horsch and Poole, 1990) and many more have since emerged (De Raedt et al., 2008; Getoor and Taskar, 2007). In these models, random variables are parameterized by individuals belonging to a population. Even for very simple first-order models, inference at the propositional level—that is, inference that explicitly considers every individual—is intractable. The idea behind *lifted inference* is to carry out as much inference as possible without propositionalizing. An exact lifted inference procedure for first-order probabilistic directed models was originally proposed by Poole (2003). It was later extended to a broader range of problems by de Salvo Braz et al. (2007). Further work by Milch et al. (2008) expanded the scope of lifted inference and resulted in the C-FOVE algorithm, which is currently the state of the art in exact lifted inference.

First-order models typically contain constraints on the parameters (logical variables typed with populations). Constraints are important for capturing knowledge regarding particular individuals. In Poole (2003), each constraint is processed only when necessary to continue probabilistic inference. We call this approach *splitting as needed*. Conversely, in de Salvo Braz et al. (2007) all constraints are processed at the start of the inference (this procedure is called *shattering*), and at every point at which a new constraint arises. Both approaches need to use constraint processing to count the number of solutions to constraint satisfaction problems that arise during the probabilistic inference. Milch et al. (2008) adopt the shattering procedure, and avoid the need to use a constraint solver by requiring that the constraints be written in *normal form*.

The impact of constraint processing on computational efficiency of lifted inference has been largely overlooked. In this paper we address this issue and compare the approaches to constraint processing listed above, both theoretically and empirically. We show that, in the worst case, shattering may have exponentially worse space and time complexity (in the number of parameters) than splitting as needed. Moreover, writing the constraints in normal form can lead to computational costs with a complexity that is even worse than exponential. Experiments confirm our theoretical results and stress the importance of informed constraint processing in lifted inference.

We introduce key concepts and notation in Section 2 and give an overview of constraint processing during lifted inference in Section 3. In Section 4 we discuss how a specialized #CSP solver can be used during lifted inference. Theoretical results are presented in Section 5. Section 6 contains results of experiments.

## 2　PRELIMINARIES

In this section we introduce a definition of *parameterized random variables*, which are essential components of first-order probabilistic models. We also define *parfactors* (Poole, 2003), which are data structures used during lifted inference.

## 2.1 PARAMETERIZED RANDOM VARIABLES

If $\mathcal{S}$ is a set, we denote by $|\mathcal{S}|$ the size of the set $\mathcal{S}$.

A *population* is a set of *individuals*. A population corresponds to a domain in logic.

A *parameter* corresponds to a logical variable and is typed with a population. Given parameter $X$, we denote its population by $\mathcal{D}(X)$. Given a set of constraints $\mathcal{C}$, we denote a set of individuals from $\mathcal{D}(X)$ that satisfy constraints in $\mathcal{C}$ by $\mathcal{D}(X) : \mathcal{C}$.

A *substitution* is of the form $\{X_1/t_1, \ldots, X_k/t_k\}$, where the $X_i$ are distinct parameters, and each *term* $t_i$ is a parameter typed with a population or a constant denoting an individual from a population. A *ground substitution* is a substitution, where each $t_i$ is a constant.

A *parameterized random variable* is of the form $f(t_1, \ldots, t_k)$, where $f$ is a functor (either a function symbol or a predicate symbol) and $t_i$ are terms. Each functor has a set of values called the *range* of the functor. We denote the range of the functor $f$ by $range(f)$. A parameterized random variable $f(t_1, \ldots, t_k)$ represents a set of random variables, one for each possible ground substitution to all of its parameters. The range of the functor of the parameterized random variable is the domain of random variables represented by the parameterized random variable.

Let **v** denote an *assignment of values* to random variables; **v** is a function that takes a random variable and returns its value. We extend **v** to also work on parameterized random variables, where we assume that free parameters are universally quantified.

**Example 1.** Let $A$ and $B$ be parameters typed with a population $\mathcal{D}(A) = \mathcal{D}(B) = \{x_1, \ldots, x_n\}$. Let $h$ be a functor with range $\{true, false\}$. Then $h(A,B)$ is a parameterized random variable. It represents a set of $n^2$ random variables with domains $\{true, false\}$, one for each ground substitution $\{A/x_1, B/x_1\}, \{A/x_1, B/x_2\}, \ldots, \{A/x_n, B/x_n\}$. A parameterized random variable $h(x_1, B)$ represents a set of $n$ random variables with domains $\{true, false\}$, one for each ground substitution $\{B/x_1\}, \ldots, \{B/x_n\}$. Let **v** be an assignment of values to random variables. If $\mathbf{v}(h(x_1, B))$ equals *true*, each of the random variables represented by $h(x_1, B)$, namely $h(x_1, x_1), \ldots, h(x_1, x_n)$, is assigned the value *true* by **v**.

## 2.2 PARAMETRIC FACTORS

A *factor* on a set of random variables represents a function that, given an assignment of a value to each random variable from the set, returns a real number. Factors are used in the variable elimination algorithm (Zhang and Poole, 1994) to store initial conditional probabilities and intermediate results of computation during probabilistic inference in graphical models. Operations on factors include mul-tiplication of factors and summing out random variables from a factor.

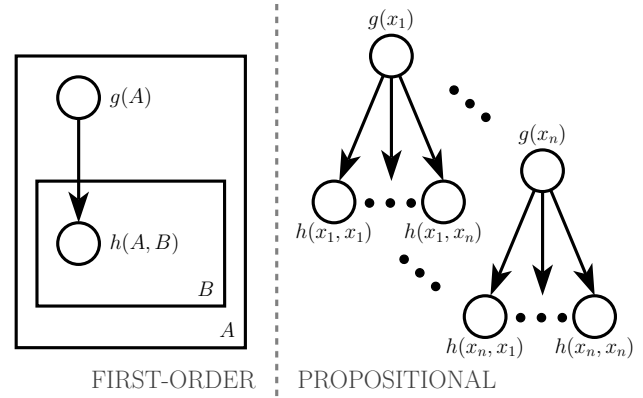

Figure 1: A parameterized belief network and its equivalent belief network.

Let **v** be an assignment of values to random variables and let $\mathcal{F}$ be a factor on a set of random variables $\mathcal{S}$. We extend **v** to factors and denote by $\mathbf{v}(\mathcal{F})$ the value of the factor $\mathcal{F}$ given **v**. If **v** does not assign values to all of the variables in $\mathcal{S}$, $\mathbf{v}(\mathcal{F})$ denotes a factor on other variables.

A *parametric factor* or *parfactor* is a triple $\langle \mathcal{C}, \mathcal{V}, \mathcal{F} \rangle$ where $\mathcal{C}$ is a set of inequality constraints on parameters, $\mathcal{V}$ is a set of parameterized random variables and $\mathcal{F}$ is a factor from the Cartesian product of ranges of parameterized random variables in $\mathcal{V}$ to the reals.

A parfactor $\langle \mathcal{C}, \mathcal{V}, \mathcal{F} \rangle$ represents a set of factors, one for each ground substitution $\mathcal{G}$ to all free parameters in $\mathcal{V}$ that satisfies the constraints in $\mathcal{C}$. Each such factor $\mathcal{F}_\mathcal{G}$ is a factor on the set of random variables obtained by applying a substitution $\mathcal{G}$. Given an assignment **v** to random variables represented by $\mathcal{V}$, $\mathbf{v}(\mathcal{F}_\mathcal{G}) = \mathbf{v}(\mathcal{F})$.

Parfactors are used to represent conditional probability distributions in directed first-order models and potentials in undirected first-order models as well as intermediate computation results during inference in first-order models.

In the next example, which extends Example 1, we use *parameterized belief networks* (PBNs) (Poole, 2003) to illustrate representational power of parfactors. The PBNs are a simple first-order directed probabilistic model, we could have used parameterized Markov networks instead (as did de Salvo Braz et al. (2007) and Milch et al. (2008)). Our discussion of constraint processing in lifted inference is not limited to PBNs, it applies to any model for which the joint distribution can be expressed as a product of parfactors.

**Example 2.** A PBN consists of a directed acyclic graph where the nodes are parameterized random variables, an assignment of a range to each functor, an assignment of a population to each parameter, and a probability distribution for each node given its parents. Consider the PBN graph presented in Figure 1 using plate notation (Buntine, 1994). Let $g$ be a functor with range $\{true, false\}$. Assume we do not have any specific knowledge about instances of $g(A)$, but we have some specific knowledge about $h(A,B)$ for case where $A = x_1$ and for case where $A \neq x_1$ and $A = B$. The probability $\mathcal{P}(g(A))$ can be represented with a parfactor $\langle \emptyset, \{g(A)\}, \mathcal{F}_g \rangle$, where $\mathcal{F}_g$ is a factor from $range(h)$ to the reals. The conditional probability $\mathcal{P}(h(A,B)|g(A))$ can be represented with a parfactor $\langle \emptyset, \{g(x_1), h(x_1,B)\}, \mathcal{F}_1 \rangle$, a parfactor $\langle \{A \neq x_1\}, \{g(A), h(A,A)\}, \mathcal{F}_2 \rangle$, and a parfactor $\langle \{A \neq x_1, A \neq B\}, \{g(A), h(A,B)\}, \mathcal{F}_3 \rangle$, where $\mathcal{F}_1$, $\mathcal{F}_2$, and $\mathcal{F}_3$ are factors from $range(g) \times range(h)$ to the reals.

Let $\mathcal{C}$ be a set of inequality constraints on parameters and $X$ be a parameter. We denote by $\mathcal{E}_X^\mathcal{C}$ the *excluded set* for $X$, that is, the set of terms $t$ such that $(X \neq t) \in \mathcal{C}$. A parfactor $\langle \mathcal{C}, \mathcal{V}, \mathcal{F}_\mathcal{F} \rangle$ is in *normal form* (Milch et al., 2008) if for each inequality $(X \neq Y) \in \mathcal{C}$, where $X$ and $Y$ are parameters, we have $\mathcal{E}_X^\mathcal{C} \setminus \{Y\} = \mathcal{E}_Y^\mathcal{C} \setminus \{X\}$.

In a normal form parfactor, for all parameters $X$ of parameterized random variables in $\mathcal{V}$, $|\mathcal{D}(X){:}\mathcal{C}| = |\mathcal{D}(X)| - |\mathcal{E}_X^\mathcal{C}|$.

**Example 3.** Consider the parfactor $\langle \{A \neq x_1, A \neq B\}, \{g(A), h(A,B)\}, \mathcal{F}_3 \rangle$ from Example 2. Let $\mathcal{C}$ denote a set of constraints from this parfactor. The set $\mathcal{C}$ contains only one inequality between parameters, namely $A \neq B$. We have $\mathcal{E}_A^\mathcal{C} = \{x_1, B\}$ and $\mathcal{E}_B^\mathcal{C} = \{A\}$. As $\mathcal{E}_A^\mathcal{C} \setminus \{B\} \neq \mathcal{E}_B^\mathcal{C} \setminus \{A\}$, the parfactor is not in normal form. Recall that $\mathcal{D}(A) = \mathcal{D}(B) = \{x_1, \ldots, x_n\}$. The size of the set $\mathcal{D}(A) : \mathcal{C}$ depends on the parameter $B$. It is equal $n-1$ when $B = x_1$ and $n-2$ when $B \neq x_1$. Other parfactors from Example 2 are in normal form as they do not contain constraints between parameters. Consider a parfactor $\langle \{X \neq Y, X \neq a, Y \neq a\}, \{e(X), f(X,Y)\}, \mathcal{F}_{ef} \rangle$, where $\mathcal{D}(X) = \mathcal{D}(Y)$ and $|\mathcal{D}(X)| = n$. Let $\mathcal{C}'$ denote a set of constraints from this parfactor. As $\mathcal{E}_X^{\mathcal{C}'} \setminus \{Y\} = \mathcal{E}_Y^{\mathcal{C}'} \setminus \{X\}$, the parfactor is in normal form and $|\mathcal{D}(X) : \mathcal{C}'| = n-2$ and $|\mathcal{D}(Y) : \mathcal{C}'| = n-2$.

# 3 LIFTED INFERENCE AND CONSTRAINT PROCESSING

In this section we give an overview of exact lifted probabilistic inference developed in (Poole, 2003), (de Salvo Braz et al., 2007), and (Milch et al., 2008) in context of constraints. For more details on other aspects of lifted inference we refer the reader to the above papers.

Let $\Phi$ be a set of parfactors. Let $\mathcal{J}(\Phi)$ denote a factor equal to the product of all factors represented by elements of $\Phi$. Let $\mathbf{U}$ be the set of all random variables represented by parameterized random variables present in parfactors in $\Phi$. Let $\mathbf{Q}$ be a subset of $\mathbf{U}$. The *marginal of $\mathcal{J}(\Phi)$ on $\mathbf{Q}$*, denoted $\mathcal{J}_\mathbf{Q}(\Phi)$, is defined as $\mathcal{J}_\mathbf{Q}(\Phi) = \sum_{\mathbf{U} \setminus \mathbf{Q}} \mathcal{J}(\Phi)$.

Given $\Phi$ and $\mathbf{Q}$, the lifted inference procedure computes the marginal $\mathcal{J}_\mathbf{Q}(\Phi)$ by summing out random variables from $\mathbf{Q}$, where possible in a lifted manner. Evidence can be handled by adding to $\Phi$ additional parfactors on observed random variables.

Before a (ground) random variable can be summed out, a number of conditions must be satisfied. One is that a random variable can be summed out from a parfactor in $\Phi$ only if there are no other parfactors in $\Phi$ involving this random variable. To satisfy this condition, the inference procedure may need to multiply parfactors prior to summing out.

Multiplication has a condition of its own: two parfactors $\langle \mathcal{C}_1, \mathcal{V}_1, \mathcal{F}_1 \rangle$ and $\langle \mathcal{C}_2, \mathcal{V}_2, \mathcal{F}_2 \rangle$ can be multiplied only if for each parameterized random variable from $\mathcal{V}_1$ and for each parameterized random variable from $\mathcal{V}_2$, the sets of random variables represented by these two parameterized random variables in respective parfactors are identical or disjoint. This condition is trivially satisfied for parameterized random variables with different functors.

**Example 4.** Consider the PBN from Figure 1 and set $\Phi$ containing parfactors introduced in Example 2. Assume that we want to compute the marginal of $\mathcal{J}(\Phi)$ on instances of $h(A,B)$, where $A \neq x_1$ and $A \neq B$. We need to sum out random variables represented by $g(A)$ from parfactor $\langle \{A \neq x_1, A \neq B\}, \{g(A), h(A,B)\}, \mathcal{F}_3 \rangle$, but as they are also among random variables represented by $g(A)$ in parfactor $\langle \emptyset, g(A), \mathcal{F}_g \rangle$, we have to first multiply these two parfactors. Sets of random variables represented by $g(A)$ in these two parfactors are not disjoint and are not identical and the precondition for multiplication is not satisfied.

## 3.1 SPLITTING

The precondition for parfactor multiplication may be satisfied through *splitting* parfactors on substitutions.

Let $\Phi$ be a set of parfactors. Let $pf = \langle \mathcal{C}, \mathcal{V}, \mathcal{F}_\mathcal{F} \rangle \in \Phi$. Let $\{X/t\}$ be a substitution such that $(X \neq t) \notin \mathcal{C}$ and term $t$ is a constant such that $t \in \mathcal{D}(X)$, or a parameter such that $\mathcal{D}(t) = \mathcal{D}(X)$. A *split* of $pf$ on $\{X/t\}$ results in two parfactors: $pf[X/t]$ that is a parfactor $pf$ with all occurrences of $X$ replaced by term $t$, and a parfactor $pf_r = \langle \mathcal{C} \cup \{X \neq t\}, \mathcal{V}, \mathcal{F}_\mathcal{F} \rangle$. We have $\mathcal{J}(\Phi) = \mathcal{J}(\Phi \setminus \{pf\} \cup \{pf[X/t], pf_r\})$. We call $pf_r$ a *residual* parfactor.

Given two parfactors that need to be multiplied, substitutions on which splitting is performed are determined by analyzing constraint sets $\mathcal{C}$ and sets of parameterized random variables $\mathcal{V}$ in these parfactors.

**Example 5.** Let us continue Example 4. A split of $\langle \emptyset, \{g(A)\}, \mathcal{F}_g \rangle$ on $\{A/x_1\}$ results in $\langle \emptyset, \{g(x_1)\}, \mathcal{F}_g \rangle$ and residual $\langle \{A \neq x_1\}, \{g(A)\}, \mathcal{F}_g \rangle$. The first parfactor can be ignored because it is not relevant to the query, while the residual needs to be multiplied by a parfactor $\langle \{A \neq x_1, A \neq B\}, \{g(A), h(A,B)\}, \mathcal{F}_3 \rangle$. The precondition for multiplication is now satisfied as $g(A)$ represents the same set of random variables in both parfactors.

## 3.2 MULTIPLICATION

Once the precondition for parfactor multiplication is satisfied, multiplication can be performed in a lifted manner. This means that, although parfactors participating in a multiplication as well as their product represent multiple factors, the computational cost of parfactor multiplication is limited to the cost of multiplying two factors. The only additional requirement is that the lifted inference procedure needs to know how many factors each parfactor involved in the multiplication represents and how many factors their product will represent. These numbers can be different because the product parfactor might involve more parameters than a parfactor participating in the multiplication. In such a case, a correction to values of a factor inside appropriate parfactors participating in multiplication is necessary. Detailed description of this correction is beyond the scope of this paper. For more information see Example 6 below and a discussion of the *fusion* operation in de Salvo Braz et al. (2007). For our purpose, the key point is that the lifted inference procedure needs to compute the number of factors represented by a parfactor.

Given a parfactor $\langle \mathcal{C}, \mathcal{V}, \mathcal{F} \rangle$, the number of factors it represents is equal to the number of solutions to the constraint satisfaction problem formed by constraints in $\mathcal{C}$. This counting problem is written as #CSP (Dechter, 2003). If a parfactor is in normal form, each connected component of the underlying constraint graph is fully connected (see Proposition 1) and it is easy to compute the number of factors represented by the considered parfactor. If a parfactor is not in normal form, a #CSP solver is necessary to compute the number of factors the parfactor represents.

**Example 6.** In Example 5 the parfactor $\langle \{A \neq x_1\}, \{g(A)\}, \mathcal{F}_g \rangle$ represents $n-1$ factors, and needs to be multiplied by a parfactor $\langle \{A \neq x_1, A \neq B\}, \{g(A), h(A,B)\}, \mathcal{F}_3 \rangle$, which represents $(n-1)^2$ factors. Their product $pf_* = \langle A \neq x_1, A \neq B, \{g(A), h(A,B)\}, \mathcal{F}_* \rangle$, where $\mathcal{F}_*$ is a factor from $range(g) \times range(h)$ to the reals, represents $(n-1)^2$ factors. Let $\mathbf{v}$ be an assignment of values to $g(A)$ and $h(A,B)$. We have $\mathbf{v}(\mathcal{F}_*) = \mathbf{v}(\mathcal{F}_g)^{(n-1)/(n-1)^2} \mathbf{v}(\mathcal{F}_3)$. Now we can sum out $g(A)$ from $pf_*$. The result needs to be represented with a *counting formula* (Milch et al., 2008), which is outside of the scope of this paper. What is important for the paper is that a parfactor involving counting formulas needs to be in normal form. Since $pf_*$ is not in normal form, we first split it on substitution $\{B/x_1\}$ and then sum out $g(A)$ from the two parfactors obtained through splitting.

## 3.3 SUMMING OUT

During lifted summing out, a parameterized random variable is summed out from a parfactor $\langle \mathcal{C}, \mathcal{V}, \mathcal{F}_\mathcal{F} \rangle$, which means that a random variable is eliminated from each factor represented by the parfactor in one inference step. Lifted inference will perform summing out only once on the factor $\mathcal{F}$. If some parameters only appear in the parameterized variable that is being eliminated, the resulting parfactor will represent fewer factors than the original one. As in the case of parfactor multiplication, the inference procedure needs to compensate for this difference. It needs to compute the size of the set $\mathcal{X} = (\mathcal{D}(X_1) \times \cdots \times \mathcal{D}(X_k)) : \mathcal{C}$, where $X_1, \ldots, X_k$ are parameters that will disappear from the parfactor. This number tells us how many times fewer factors the result of summing out represents compared to the original parfactor.

If the parfactor is in normal form, $|\mathcal{X}|$ does not depend on values of parameters remaining in the parfactor after summing out. The problem reduces to #CSP and, as we mentioned in Section 3.2, it is easy to solve. If the parfactor is not in normal form, $|\mathcal{X}|$ may depend on values of remaining parameters and a #CSP solver is necessary to compute all the sizes of the set $\mathcal{X}$ conditioned on values of parameters remaining in the parfactor.

**Example 7.** Assume that we want to sum out $f(X,Y)$ from a parfactor $\langle \{X \neq Y, Y \neq a\}, \{e(X), f(X,Y)\}, \mathcal{F}_{ef} \rangle$, where $\mathcal{F}_{ef}$ is a factor from $range(e) \times range(f)$ to the reals. Let $\mathcal{D}(X) = \mathcal{D}(Y)$ and $|\mathcal{D}(X)| = n$. The parfactor represents $(n-1)^2$ factors. Note that the parfactor is not in normal form and $|\mathcal{D}(Y) : \{X \neq Y, Y \neq a\}|$ equals $n-1$ if $X = a$ and $n-2$ if $X \neq a$. A #CSP solver could compute these numbers for us (see Example 9). After $f(X,Y)$ is summed out, $Y$ is no longer among parameters of random variables and $X$ remains the sole parameter. To represent the result of summation we need two parfactors: $\langle \emptyset, e(a), \mathcal{F}_{e1} \rangle$ and $\langle \{X \neq a\}, e(X), \mathcal{F}_{e2} \rangle$, where $\mathcal{F}_{e1}$ and $\mathcal{F}_{e2}$ are factors from $range(e)$ to the reals. Let $y \in range(e)$, then $\mathcal{F}_{e1}(y) = (\sum_{z \in range(f)} \mathcal{F}_{ef}(y,z))^{n-1}$ and $\mathcal{F}_{e2}(y) = (\sum_{z \in range(f)} \mathcal{F}_{ef}(y,z))^{n-2}$.

## 3.4 PROPOSITIONALIZATION

During inference in first-order probabilistic models it may happen that none of lifted operations (including operations that are not described in this paper) can be applied. In such a situation the inference procedure substitutes appropriate parameterized random variables with random variables represented by them. This may be achieved through splitting as we demonstrate in an example below. Afterward, inference is performed, at least partially, at the propositional level. As it has a negative impact on the efficiency of inference, propositionalization is avoided as much as possible during inference in first-order models.

**Example 8.** Consider a parfactor $\langle \emptyset, \{g(A)\}, \mathcal{F}_g \rangle$ from Example 2. Assume we need to propositionalize $g(A)$. Recall that $\mathcal{D}(A) = \{x_1, \ldots, x_n\}$. Propositionalization results in a set of parfactors $\langle \emptyset, \{g(x_1)\}, \mathcal{F}_g \rangle$, $\langle \emptyset, \{g(x_2)\}, \mathcal{F}_g \rangle$, $\ldots$, $\langle \emptyset, \{g(x_{n-1})\}, \mathcal{F}_g \rangle$, $\langle \emptyset, \{g(x_n)\}, \mathcal{F}_g \rangle$. Each parameterized random variable in the above parfactors represents just one random variable and each parfactor represents just one factor. The set could be produced by a sequence of splits.

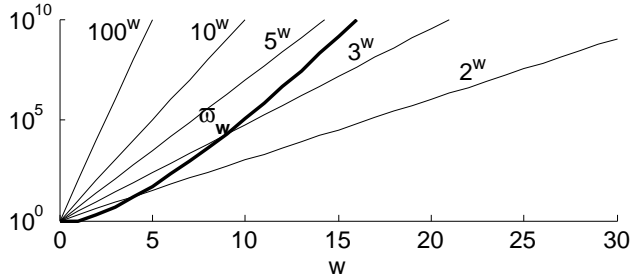

Figure 2: Comparison of $\varpi_w$ and exponential functions.

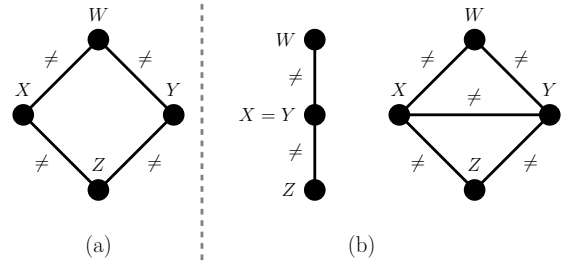

Figure 3: Constraint graph with a cycle (a) and the two cases that need to be analyzed (b).

The above informal overview of lifted inference, together with simple examples, shows that constraint processing is an integral, important part of lifted inference.

## 4  #CSP SOLVER AND LIFTED INFERENCE

In Section 3, we showed when a #CSP solver can be used during lifted probabilistic inference. A solver that enumerates all individuals from domains of parameters that form a CSP would contradict the core idea behind lifted inference, that is, performing probabilistic inference without explicitly considering every individual. In our experiments presented in Section 6 we used a solver (Kisyński and Poole, 2006) that addresses the above concern. It is a lifted solver based on the variable elimination algorithm for solving #CSP of Dechter (2003) and is optimized to handle problems that contain only inequality constraints. It is not possible to describe the algorithm in detail in this paper, but below we provide some intuition behind the solver.

First, we need to introduce a concept of a *set partition*. A partition of a set $S$ is a collection $\{B_1, \ldots, B_w\}$ of nonempty, pairwise disjoint subsets of $S$ such that $S = \bigcup_{i=1}^{w} B_i$. The sets $B_i$ are called *blocks* of the partition. Set partitions are intimately connected to equality. For any consistent set of equality assertions on parameters, there is a partition in which the parameters that are equal are in the same block, and the parameters that are not equal are in different blocks. If we consider a semantic mapping from parameters to individuals in the world, the inverse of this mapping, where two parameters that map to the same individual are in the same block, forms a partition of the parameters. The number of partitions of the set of size $w$ is equal to the $w$-th *Bell number* $\varpi_w$. Bell numbers grow faster than any exponential function (see Lovász (2003)), but for small $w$'s they stay much smaller than exponential functions with a moderate base (see Figure 2).

Consider a set of constraints $\{W \neq X, X \neq Y, X \neq Z\}$, where all parameters have the same population of size $n$. The underlying constraint graph has a tree structure, which allows us to immediately solve the corresponding #CSP: we can assign the value to $W$ in $n$ ways, and are left with $n-1$ possible values for $X$, and $n-1$ possible values for $Y$ and $Z$. Hence, there are $n(n-1)^3$ solutions to this CSP instance.

Consider a set of constraints $\{W \neq X, W \neq Y, X \neq Z, Y \neq Z\}$, where all parameters have the same population of size $n$. The underlying graph has a cycle (see Figure 3 (a)), which makes the corresponding #CSP more difficult to solve than in the previous example. We can assign the value to $W$ in $n$ ways, and are left with $n-1$ possible values for $X$ and $Y$. For $Z$ we need to consider two cases: $X = Y$ and $X \neq Y$ (see Figure 3 (b)). In the $X = Y$ case, $Z$ can take $n-1$ values, while in the $X \neq Y$ case, $Z$ can have $n-2$ different values. Hence, the number of solutions to this CSP instance is $n(n-1)^2 + n(n-1)(n-2)^2$. The first case corresponds to a partition $\{\{X,Y\}\}$ of the set of parameters $\{X,Y\}$, while the second case corresponds to a partition $\{\{X\},\{Y\}\}$ of this set.

In general, to perform this kind of reasoning, we need to triangulate the constraint graph. This can be naturally achieved with a variable elimination algorithm. Each new edge adds two cases, one in which the edge corresponds to the equality constraint and one in which it corresponds to the inequality constraint. Some cases are inconsistent and can be ignored. When we have to analyze a fully connected subgraph of $w$ new edges, we need to consider $\varpi_w$ cases. This is because each such case corresponds to a partition of the parameters from the subgraph; those parameters in the same block of the partition are equal, and parameters in different blocks are not equal. The number of such partitions is equal to $\varpi_w$. The lifted #CSP solver analyzes $\varpi_w$ partitions of parameters, rather than $n^w$ ground substitutions of individuals. Since we do not care about empty partitions, we will never have to consider more partitions than there are ground substitutions. As $w$ corresponds to the induced width of a constraint graph (which we do not expect to be large) and $n$ corresponds to the population size (potentially large), the difference between $\varpi_w$ and $n^w$ can be very big (see Figure 2).

In practice, parameters can be typed with different populations (from the very beginning as well as because of unary constraints). In such a situation, we can apply the above reasoning to any set of individuals that are indistinguishable as far as counting is concerned. For example, the intersection of all populations is a set of individuals for which we only need the size; there is no point in reasoning about each individual separately. Similarly, the elements from the population of a parameter that do not belong to the population of any other parameter can be grouped and treated together. In general, any individuals that are in the populations of the same group of parameters can be treated identically; all we need is to know how many there are.

**Example 9.** In Example 7 we need to know the number $|\mathcal{D}(Y) : \{X \neq Y, Y \neq a\}|$, where $\mathcal{D}(X) = \mathcal{D}(Y)$ and $|\mathcal{D}(X)| = n$. Let $\mathbf{a}_1$ denote set $\{a\}$ and $\mathbf{a}_2$ denote set $\mathcal{D}(X) \setminus \{a\}$. The following factor has value 1 for substitutions to parameters $X,Y$ that are solutions to the above CSP and 0 otherwise:

| $X$ | $Y$ | Partition(s) | |
|---|---|---|---|
| $\mathbf{a}_1$ | $\mathbf{a}_2$ | $\{\{X\}\}\{\{Y\}\}$ | 1 |
| $\mathbf{a}_2$ | $\mathbf{a}_2$ | $\{\{X,Y\}\}$ | 0 |
| $\mathbf{a}_2$ | $\mathbf{a}_2$ | $\{\{X\},\{Y\}\}$ | 1 |

After we eliminate $Y$ from the above factor we obtain:

| $X$ | Partition(s) | |
|---|---|---|
| $\mathbf{a}_1$ | $\{\{X\}\}$ | $n-1$ |
| $\mathbf{a}_2$ | $\{\{X\}\}$ | $n-2$ |

Numbers $n-1$ and $n-2$ are obtained through analysis of partitions of $X$ and $Y$ present in the original factor and knowledge that $\mathbf{a}_1$ represents 1 individual and $\mathbf{a}_2$ represents $n-1$ individuals. From the second factor we can infer that $|\mathcal{D}(Y) : \{X \neq Y, Y \neq a\}|$ equals $n-1$ if $X = a$ and $n-2$ if $X \neq a$.

If we assume that all populations of parameters forming a #CSP are sorted according to the same (arbitrary) ordering, sets of indistinguishable individuals can be generated through a single sweep of the populations. Each such set can be represented by listing all of its elements or by listing all elements from the corresponding population that do not belong to it. For each set we choose a more compact representation.

The answer from the solver needs to be translated to sets of substitutions and constraints accompanying each computed value. Standard combinatorial enumeration algorithms can do this task.

## 5 THEORETICAL RESULTS

In this section we discuss consequences of different approaches to constraint processing in lifted inference.

### 5.1 SPLITTING AS NEEDED VS. SHATTERING

Poole (2003) proposed a scheme in which splitting is performed "as needed" through the process of inference when two parfactors are about to be multiplied and the precondition for multiplication is not satisfied.

An alternative, called *shattering*, was proposed by de Salvo Braz et al. (2007). They perform splitting at the beginning of the inference by doing all the splits that are required to ensure that for any two parameterized random variables present in considered parfactors the sets of random variables represented by them are either identical or disjoint.[1] Shattering was also used in Milch et al. (2008).

Shattering simplifies design and implementation of lifted inference procedures, in particular, construction of elimination ordering heuristics. Unfortunately, as we show in Theorem 1, it may lead to creation of large number of parfactors that would not be created by following the splitting as needed approach.

**Theorem 1.** *Let $\Phi$ be a set of parfactors. Let $\mathbf{Q}$ be a subset of the set of all random variables represented by parameterized random variables present in parfactors in $\Phi$. Assume we want to compute the marginal $\mathcal{J}_{\mathbf{Q}}(\Phi)$. Then:*

*(i) if neither of the algorithms performs propositionalization, then every split on substitution $\{X/t\}$, where $t$ is a constant, performed by lifted inference with splitting as needed is also performed by lifted inference with shattering (subject to a renaming of parameters);*

*(ii) lifted inference with shattering might create exponentially more (in the maximum number of parameters in a parfactor) parfactors than lifted inference with splitting as needed.*

*Proof.* We present a sketch of a proof of the first statement and a constructive proof of the second statement.

*(i)* Assume that lifted inference with splitting as needed performs a split. We can track back the cause of this to the initial set of parfactors $\Phi$. Further analysis shows that shattering the set $\Phi$ would also involve this split.

*(ii)* Consider the following set of parfactors:

$$\Phi = \{\langle \emptyset, \{g_Q(), g_1(X_1, X_2, \ldots, X_k),$$
$$g_2(X_2, X_3, \ldots, X_k),$$
$$\ldots,$$
$$g_k(X_k))\}, \mathcal{F}_0 \rangle, \quad [0]$$
$$\langle \emptyset, \{g_1(a, X_2, \ldots, X_k)\}, \mathcal{F}_1 \rangle, \quad [1]$$
$$\langle \emptyset, \{g_2(a, X_3, \ldots, X_k)\}, \mathcal{F}_2 \rangle, \quad [2]$$
$$\ldots,$$
$$\langle \emptyset, \{g_{k-1}(a, X_k)\}, \mathcal{F}_{k-1} \rangle, \quad [k-1]$$
$$\langle \emptyset, \{g_k(a)\}, \mathcal{F}_k \rangle, \quad [k]$$

and let $\mathbf{Q}$ be $g_Q()$.

---
[1] Shattering might also be necessary in the middle of inference if propositionalization has been performed.

For $i = 1, \ldots, k$, a set of random variables represented by a parameterized random variable $g_i(X_i, \ldots, X_k)$ in a parfactor [0] is a proper superset of a set of random variables represented by a parameterized random variable $g_i(a, X_{i+1}, \ldots, X_k)$ in a parfactor [i]. Therefore lifted inference with shattering needs to perform several splits. Since the order of splits during shattering does not matter here, assume that the first operation is a split of the parfactor [0] on a substitution $\{X_1/a\}$ which creates a parfactor

$$\langle \emptyset, \{g_Q(), g_1(a, X_2, \ldots, X_k), g_2(X_2, X_3, \ldots, X_k), \ldots, g_k(X_k))\}, \mathcal{F}_0 \rangle \qquad [k+1]$$

and a residual parfactor

$$\langle \{X_1 \neq a\}, \{g_Q(), g_1(X_1, X_2, \ldots, X_k), g_2(X_2, X_3, \ldots, X_k), \ldots, g_k(X_k))\}, \mathcal{F}_0 \rangle. \qquad [k+2]$$

In both newly created parfactors, for $i = 2, \ldots, k$, a set of random variables represented by a parameterized random variable $g_i(X_i, \ldots, X_k)$ is a proper superset of a set of random variables represented by a parameterized random variable $g_i(a, X_{i+1}, \ldots, X_k)$ in a parfactor [i] and shattering proceeds with further splits of both parfactors. Assume that in next step parfactors [k + 1] and [k + 2] are split on a substitution $\{X_2/a\}$. The splits result in four new parfactors. The result of the split of the parfactor [k + 1] on $\{X_2/a\}$ contains a parameterized random variable $g_1(a, a, \ldots, X_k)$ and a parfactor [1] needs to be split on a substitution $\{X_2/a\}$. The shattering process continues following a scheme described above. It terminates after $2^{k+1} - k - 2$ splits and results in $2^{k+1} - 1$ parfactors (each original parfactor [i], $i = 0, \ldots, k$, is shattered into $2^{k-i}$ parfactors). Assume that lifted inference proceeds with an elimination ordering $g_1, \ldots, g_k$ (this elimination ordering does not introduce counting formulas, other do). To compute the marginal $\mathcal{J}_{g_Q()}(\Phi)$, $2^k$ lifted multiplications and $2^{k+1} - 2$ lifted summations are performed.

Consider lifted inference with splitting as needed. Assume it follows an elimination ordering $g_1, \ldots, g_k$. A set of random variables represented by a parameterized random variable $g_1(X_1, \ldots, X_k)$ in a parfactor [0] is a proper superset of a set of random variables represented by a parameterized random variable $g_1(a, X_2, \ldots, X_k)$ in a parfactor [1] and the parfactor [0] is split on a substitution $\{X_1/a\}$. The split results in parfactors identical to the parfactors [k + 1] and [k + 2] from the description of shattering above. The parfactor [k + 1] is multiplied by the parfactor [1] and all instances of $g_1(a, X_1, \ldots, X_k)$ are summed out from their product while all instances of $g_1(X_1, X_2, \ldots, X_k)$ (subject to a constraint $X_1 \neq a$) are summed out from the parfactor [k + 2]. The summations create two parfactors:

$$\langle \emptyset, \{g_Q(), g_2(X_2, X_3, \ldots, X_k), \ldots, g_k(X_k))\}, \mathcal{F}_{\mathcal{F}_{k+3}} \rangle, \qquad [k+3]$$
$$\langle \emptyset, \{g_Q(), g_2(X_2, X_3, \ldots, X_k), \ldots, g_k(X_k))\}, \mathcal{F}_{\mathcal{F}_{k+4}} \rangle. \qquad [k+4]$$

Instances of $g_2$ are eliminated next. Parfactors [k + 3] and [k + 4] are split on a substitution $\{X_2/a\}$, the results of the splits and a parfactor [2] are multiplied together and the residual parfactors are multiplied together. Then, all instances of $g_2(a, X_3, \ldots, X_k)$ are summed out from the first product product while all instances of $g_2(X_2, \ldots, X_k)$ (subject to a constraint $X_2 \neq a$) are summed out from the second product. The elimination of $g_3, \ldots, g_k$ looks the same as for $g_2$. In total, $2k - 1$ splits, $3k - 2$ lifted multiplications and $2k$ lifted summations are performed. At any moment, the maximum number of parfactors is $k + 3$. □

The above theorem shows that shattering approach is never better and sometimes worse than splitting as needed. It is worth pointing out that splitting as needed approach complicates the design of an elimination ordering heuristic.

### 5.2 NORMAL FORM PARFACTORS VS. #CSP SOLVER

Normal form parfactors were introduced by Milch et al. (2008) in the context of counting formulas. Counting formulas are parameterized random variables that let us compactly represent a special form of probabilistic dependencies between instances of a parameterized random variable. Milch et al. (2008) require all parfactors to be in normal form to eliminate the need to use a separate constraint solver to solve #CSP. The requirement is enforced by splitting parfactors that are not in normal form on appropriate substitutions. While parfactors that involve counting formulas must be in normal form, that is not necessary for parfactors without counting formulas. It might actually be quite expensive as we show in this section.

**Proposition 1.** *Let $\langle \mathcal{C}, \mathcal{V}, \mathcal{F} \rangle$ be a parfactor in normal form. Then each connected component of the constraint graph corresponding to $\mathcal{C}$ is fully connected.*

*Proof.* Proposition 1 is trivially true for components with one or two parameters. Let us consider a connected component with more than two parameters. Suppose, contrary to our claim, that there are two parameters $X$ and $Y$ with no edge between them. Since the component is connected, there exists a path $X, Z_1, Z_2, \ldots, Z_m, Y$. As $\mathcal{C}$ is in normal form, $\mathcal{E}_{Z_i}^{\mathcal{C}} \setminus \{Z_{i+1}\} = \mathcal{E}_{Z_{i+1}}^{\mathcal{C}} \setminus \{Z_i\}$, $i = 1, \ldots, m-1$ and $\mathcal{E}_{Z_m}^{\mathcal{C}} \setminus \{Y\} = \mathcal{E}_Y^{\mathcal{C}} \setminus \{Z_m\}$. We have $X \in \mathcal{E}_{Z_1}^{\mathcal{C}}$, and consequently $X \in \mathcal{E}_Y^{\mathcal{C}}$. This contradicts our assumption that there is no edge between $X$ and $Y$. □

While the above property simplifies solving #CSP for a set of constraints from a parfactor in normal form it also has negative consequences. If a parfactor is not in normal form, conversion to normal from might require several splits. For example we need three splits to convert a parfactor with the set of constraints shown in Figure 3 (a) to a set of four parfactors in normal form. The resulting sets of constraints are presented in Figure 4. If the underlying graph is sparse, conversion might be very expensive as we show in the example below.

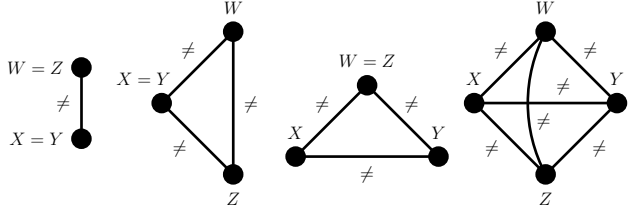

Figure 4: Constraint graphs obtained through a conversion to normal form.

**Example 10.** Consider a parfactor $\langle \{X_0 \neq a, X_0 \neq X_1, \ldots, X_0 \neq X_k\}, \{g_0(X_0), g_1(X_1), \ldots, g_n(X_k)\}, \mathcal{F}\rangle$, where $\mathcal{D}(X_0) = \mathcal{D}(X_1) = \cdots = \mathcal{D}(X_k)$. Let $\mathcal{C}$ denote a set of constraints from this parfactor. We have $\mathcal{E}^\mathcal{C}_{X_0} = \{a, X_1, \ldots, X_k\}$ and $\mathcal{E}^\mathcal{C}_{X_i} = \{X_0\}$, $i = 1, \ldots, k$. The parfactor is not in normal form because $\mathcal{E}^\mathcal{C}_{X_0} \setminus \{X_i\} \neq \mathcal{E}^\mathcal{C}_{X_i} \setminus \{X_0\}$, $i = 1, \ldots, k$. As a result the size of the set $\mathcal{D}(X_0) : \mathcal{C}$ depends on other parameters in the parfactor. For instance, it differs for $X_1 = a$ and $X_1 \neq a$ or for $X_1 = X_2$ and $X_1 \neq X_2$. A conversion of the considered parfactor to set of parfactors in normal form involves $2^k - 1$ splits on substitutions of the form $\{X_i/a\}$, $1 \leq i \leq k$ and $\sum_{i=2}^k \binom{k}{i}(\varpi_i - 1)$ splits on substitutions of the form $\{X_i/X_j\}$, $1 \leq i, j \leq k$. It creates $\sum_{i=0}^k \binom{k}{i}\varpi_i$ parfactors in normal form. In Example 11 we analyze how this conversion affects parfactor multiplication compared to the use of a #CSP solver.

From the above example we can clearly see that the cost of converting a parfactor to normal form can be worse than exponential. Moreover, converting parfactors to normal form may be very inefficient when analyzed in context of parfactor multiplication (see Section 5.2.1) or summing out a parameterized random variable from a parfactor (see Section 5.2.1). Our empirical tests (see Section 6.2) confirm this observation.

Note that splitting as needed can be used together with a #CSP solver (Poole, 2003), or with normal form parfactors. Shattering can be used with a #CSP solver (de Salvo Braz et al., 2007) or with normal form parfactors (Milch et al., 2008). The cost of converting parfactors to normal form might be amplified if it is combined with shattering.

### 5.2.1 Multiplication

In the example below we demonstrate how the normal form requirement might lead to a lot of, otherwise unnecessary, parfactor multiplications.

**Example 11.** Assume we would like to multiply the parfactor from Example 10 by a parfactor $pf = \langle \emptyset, \{g_1(X_1)\}, \mathcal{F}_1\rangle$. First, let us consider how it is done with a #CSP solver. A #CSP solver computes the number of factors the parfactor from Example 10 represents, $(|\mathcal{D}(X_0)| - 1)^{k+1}$. Next the solver computes the number of factors represented by the parfactor $pf$, which is trivially $|\mathcal{D}(X_1)|$. A correction is applied to values of the factor $\mathcal{F}_1$ to compensate for the difference between these two numbers. Finally the two parfactors are multiplied. The whole operation involved two calls to a #CSP solver, one correction and one parfactor multiplication. Now, let us see how it can be done without the use of #CSP solver. The first parfactor is converted to a set $\Phi$ of $\sum_{i=0}^k \binom{k}{i}\varpi_i$ parfactors in normal form, as presented in Example 10. Some of the parfactors in $\Phi$ contain a parameterized random variable $g_1(a)$, the rest contains a parameterized random variable $g_1(X)$ and a constraint $X_1 \neq a$, so the parfactor $pf$ needs to be split on a substitution $\{X_1/a\}$. The split results in a parfactor $\langle \emptyset, \{g_1(a)\}, \mathcal{F}_1\rangle$ and a residual $\langle \{X_1 \neq a\}, \{g_1(X_1)\}, \mathcal{F}_1\rangle$. Next, each parfactor from $\Phi$ is multiplied by either the result of the split or the residual. Thus $\sum_{i=0}^k \binom{k}{i}\varpi_i$ parfactor multiplications need to be performed and most of these multiplication require a correction prior to the actual parfactor multiplication.

There is an opportunity for some optimization, as factor components of parfactors multiplications for different corrections could be cached and reused instead of being recomputed. Still, even with such a caching mechanism, multiple parfactor multiplications would be performed compared to just one multiplication when a #CSP solver is used.

### 5.2.2 Summing Out

Examples 7 and 9 demonstrate how summing out a parameterized variable from a parfactor that is not in normal form can be done with a help of a #CSP solver. In the example below we show how this operation would look if we convert the parfactor to a set of parfactors in normal form which does not require a #CSP solver.

**Example 12.** Assume that we want to sum out $f(X,Y)$ from the parfactor $\langle \{X \neq Y, Y \neq a\}, \{e(X), f(X,Y)\}, \mathcal{F}_{ef}\rangle$ from the Example 7. First, we convert it to a set of parfactors in normal form by splitting on a substitutions $\{X/a\}$. We obtain two parfactors in normal form: $\langle \{Y \neq a\}, \{e(a), f(a,Y)\}, \mathcal{F}_{ef}\rangle$, which represents $n-1$ factors, and $\langle \{X \neq Y, X \neq a, Y \neq a\}, \{e(X), f(X,Y)\}, \mathcal{F}_{ef}\rangle$, which represents $(n-1)(n-2)$ factors. Next we sum out $f(a,Y)$ from the first parfactor and $f(X,Y)$ from the second parfactor. In both cases a correction will be necessary, as Y will no longer among parameters of random variables and the resulting parfactors will represent fewer factors than the original parfactors.

In general, as illustrated by Examples 7, 9 and 12, conversion to normal form and #CSP solver create the same number of parfactors. The difference is, that the first method, computes a factor component for the resulting parfactors once and then applies a different correction for each result-ing parfactor based on the answer from the #CSP solver. The second method computes the factor component multiple times, once for each resulting parfactor, but does not use a #CSP solver. As these factor components (before applying a correction) are identical, redundant computations could be eliminated by caching. We successfully adopted a caching mechanism in our empirical test (Section 6.2), but expect it to be less effective for larger problems.

As in the case of splitting as needed, it might be difficult to design an efficient elimination ordering heuristic that would work with a #CSP solver. This is because we do not known in advance how many parfactors will be obtained as a result of summing out. We need to run a #CSP solver to obtain this information.

## 6 EXPERIMENTS

We used Java implementations of tested lifted inference methods. Tests were performed on an Intel Core 2 Duo 2.66GHz processor with 1GB of memory made available to the JVM.

### 6.1 SPLITTING AS NEEDED VS. SHATTERING

In the first experiment we checked to what extent the overhead of the shattering approach can be minimized by using intensional representations and immutable objects that are shared whenever possible. We ran tests on the following set of parfactors:

$$\Phi = \{ \langle \emptyset, \{g_Q(), g_1(a)\}, \mathcal{F}_0 \rangle, \qquad [0]$$
$$\langle \{X \neq a\}, \{g_Q(), g_1(X)\}, \mathcal{F}_1 \rangle, \qquad [1]$$
$$\langle \emptyset, \{g_1(X), g_2(X)\}, \mathcal{F}_2 \rangle, \qquad [2]$$
$$\langle \emptyset, \{g_2(X), g_3(X)\}, \mathcal{F}_3 \rangle, \qquad [3]$$
$$\ldots,$$
$$\langle \emptyset, \{g_{k-1}(X), g_k(X)\}, \mathcal{F}_k \rangle, \qquad [k]$$
$$\langle \emptyset, \{g_k(X)\}, \mathcal{F}_{k+1} \rangle \} \qquad [k+1].$$

All functors had the range size 10 and we set $\mathbf{Q}$ to the instance of $g_Q()$. We computed the marginal $\mathcal{J}_\mathbf{Q}(\Phi)$. Lifted inference with shattering first performed total of $k$ splits, then proceeded with $2k+1$ multiplications and $2k$ summations regardless of the elimination ordering. Lifted inference with splitting as needed performed 1 split, $k+2$ multiplications and $k+1$ summations (for the experiment we used the best elimination ordering, that is $g_k, g_{k-1}, \ldots, g_1$).

Figure 5 shows the results of the experiment where we varied $k$ from 1 to 100. Even though lifted inference with shattering used virtually the same amount of memory as lifted inference with splitting, it was slower because it performed more arithmetic operations.

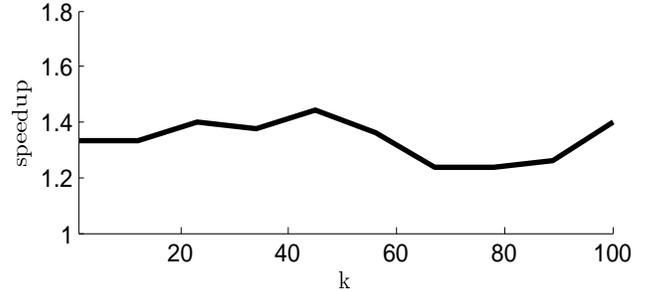

Figure 5: Speedup of splitting as needed over shattering.

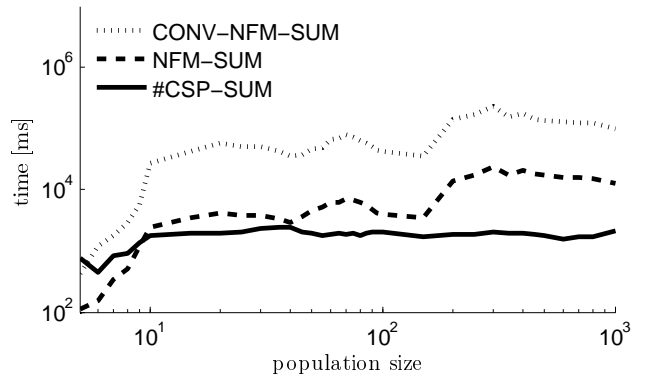

Figure 6: Summing out with and without a #CSP solver.

### 6.2 NORMAL FORM PARFACTORS VS. #CSP SOLVER

For experiment in this section we randomly generated sets of parfactors. There were up to 5 parameterized random variables in each parfactor with range sizes varying from 2 to 10. Constraints sets contained very few (and very often zero) constraints and formed sparse CSPs. Most of parfactors were in normal form, which allowed us to account for #CSP solver overhead. There were up to 10 parameters present in each parfactor. Parameters were typed with the same population. We varied the size of this population from 5 to 1000 to verify how well #CSP solver scaled for larger populations.

In this experiment we summed out a parameterized random variable from a parfactor. We compared summing out with a help of a #CSP solver (#CSP-SUM) to summing out achieved by converting a parfactor to a set of parfactors in normal form and summing out a parameterized random variable from each obtained parfactor without a #CSP solver. (We cached factor components as suggested in Section 5.2.2). For each population size we generated 100 parfactors and reported a cumulative time. For the second approach, we reported time including (CONV-NFM-SUM) and excluding (NFM-SUM) conversion to normal form. Results presented on Figure 6 show significant cost of conversion to normal form and advantage of #CSP solver for larger population sizes.

# 7 CONCLUSIONS AND FUTURE WORK

In this paper we analyzed the impact of constraint processing on the efficiency of lifted inference, and explained why we cannot ignore its role in lifted inference. We showed that a choice of constraint processing strategy has big impact on efficiency of lifted inference. In particular, we discovered that shattering (de Salvo Braz et al., 2007) is never better—and sometimes worse—than splitting as needed (Poole, 2003), and that the conversion of parfactors to normal form (Milch et al., 2008) is an expensive alternative to using a specialized #CSP solver. Although in this paper we focused on exact lifted inference, our results are applicable to approximate lifted inference. For example, see the recent work of (Singla and Domingos, 2008) that uses shattering.

It is difficult to design an elimination ordering heuristic that works well with the splitting as needed approach and a #CSP solver. We plan to address this problem in our future research.

### Acknowledgments

The authors wish to thank Brian Milch for discussing the C-FOVE algorithm with us. Peter Carbonetto, Michael Chiang and Mark Crowley provided many helpful suggestions during the preparation of the paper. This work was supported by NSERC grant to David Poole.